\documentclass[conference]{IEEEtran}
\IEEEoverridecommandlockouts
\usepackage{cite}
\usepackage{amsmath,amssymb,amsfonts}
\usepackage{lipsum}
\usepackage{multicol}
\usepackage{graphicx}
\usepackage{multirow}
\usepackage{tabularx}

\usepackage[utf8]{inputenc}
\usepackage{mathtools}
\usepackage{algorithm}
\usepackage{algpseudocode}
\usepackage{graphicx}
\usepackage{textcomp}
\makeatletter

\newcommand{\Rmnum}[1]{\expandafter\@slowromancap\romannumeral #1@}
\makeatother
\def\BibTeX{{\rm B\kern-.05em{\sc i\kern-.025em b}\kern-.08em
    T\kern-.1667em\lower.7ex\hbox{E}\kern-.125emX}}
\begin{document}

\title{Interpretable  Deep Convolutional Neural Networks via Meta-learning\\
}

\author{\IEEEauthorblockN{Xuan Liu\IEEEauthorrefmark{1},
Xiaoguang Wang\IEEEauthorrefmark{1}\IEEEauthorrefmark{2},
Stan Matwin\IEEEauthorrefmark{1}\IEEEauthorrefmark{3}}
\IEEEauthorblockA{\IEEEauthorrefmark{1}Institute
for Big Data Analytics\\
Faculty of Computer Science, Dalhousie University, Halifax, NS, Canada\\
Email: xuan.liu@dal.ca}
\IEEEauthorblockA{\IEEEauthorrefmark{2}Alibaba Group, Hangzhou, China\\
Email: xiaoguang.wxg@alibaba-inc.com }
\IEEEauthorblockA{\IEEEauthorrefmark{3}Institute of Computer Science\\
Polish Academy of Sciences, Warsaw, Poland\\
Email: stan@cs.dal.ca}}
\maketitle              

\begin{abstract}
Model interpretability is a requirement in many applications in which crucial decisions are made by users relying on a model's outputs. The recent movement for ``algorithmic fairness" also  stipulates explainability, and therefore interpretability of learning models. And yet the most successful contemporary Machine Learning approaches, the Deep Neural Networks, produce models that are highly non-interpretable. We  attempt to address this challenge by proposing a technique called CNN-INTE to interpret deep Convolutional Neural Networks (CNN) via meta-learning. In this work, we interpret a specific hidden layer of the deep CNN model on the MNIST image dataset. We use a clustering algorithm in a two-level structure to find the meta-level training data and Random Forest as base learning algorithms to generate the meta-level test data. The interpretation results are displayed visually via diagrams, which clearly indicates how a specific test instance is classified. Our method achieves global interpretation for all the test instances on the hidden layers without sacrificing the accuracy obtained by the original deep CNN model. This means our model is faithful to the original deep CNN model, which leads to reliable interpretations.  
\end{abstract}


\section{Introduction}
With the fast development of sophisticated machine learning algorithms, artificial intelligence has been gradually penetrating  a number of brand new fields with unprecedented speed.  One of the outstanding problems hampering  further progress is the \textit{interpretability} challenge.  This challenge arises when the models built by the  machine learning algorithms are to be used by humans in their decision making, particularly when such decisions are subject to legal consequences and/or  administrative audits.  For  human decision makers operating in those circumstances, to accept the  professional and legal responsibility ensuing from decisions assisted by machine learning, it is   critical to comprehend the models.  This is generally true  for areas like criminal justice, health care, terrorism detection, education system and financial markets. 

To trust the model, decision makers  need to first understand the model's behavior, and then evaluate and refine the model using their domain knowledge. Even for areas like book or movie recommendations \cite{b1} and automated aids \cite{b2}, explanations for a recommendation and an error made could increase the trust and reliance on these systems.  Furthermore, the European General Data Protection Regulation, forthcoming in June, 2018, stipulates the explainability of all automatically made decisions concerning individuals, and that includes the decisions made with or assisted by machine learning models.  Hence, there is a growing demand for interpretability of the  machine learning algorithms.

In this paper,  we define interpretability of a model as the ability to provide visual or textual presentation of the connections between input features and the output predictions. 

To realize the goal of interpretability,  there are usually two approaches. One is to design an algorithm that is inherently interpretable, while achieving competitive accuracy of a complex model. The examples are Decision Trees\cite{b3}, Decision Lists\cite{b4}, and  Decision Sets\cite{b5}, etc. The disadvantage of this approach is that there is a trade off between interpretability and accuracy: it is not easy to learn  an interpretable (so presumably simple) model expressing a complex process with a very high accuracy. The other approach which does not sacrifice accuracy takes the opposite approach: it first builds a highly accurate model without worrying about interpretabilty, and subsequently  uses  a separate set of re-representation techniques to assist the user in understanding the behavior of the algorithm.  One of the techniques could be to use the aforementioned relatively simple and interpretable algorithms to explain the behavior of a complex model and the  reasons why a given  classifier, treated as a black box, classifies a given instance in a particular way, e.g. LIME \cite{b6}, BETA \cite{b7}, TREPAN\cite{b11}.  

Deep learning methods  have been lately  very  successful in image processing and natural language processing. It could be categorized as a representation learning approach  \cite{b12}, which learns refined features that could improve a model's generalization ability. Deep learning, however, is highly non-interpretable. 

In this paper we are reporting a work in progress where  we try to  interpret  the inner mechanisms of deep learning. Our method: CNN-INTE is inspired by  \cite{b8}. We design and implement a tool that helps the user understand  how the hidden layers in a deep CNN model work to classify examples. And the results are expressed in graphs which indicate sequential separations of the true class and the hypothesis. The main contributions of our method is as follows:
\begin{itemize}
\item Compared to LIME \cite{b6} which provides \textit{local} interpretations for the entire model in specific regions of the feature space, our method provides \textit{global} interpretation for any test instances on the hidden layers  in the whole feature space.
\item Compared to models which apply inherently interpretable algorithms, e.g.  \cite{b5}, our method has the advantage of  not compromising the accuracy of the model to be interpreted. This produces more reliable interpretation. 
\item In contrary to \cite{b6} and\cite{b7} which treat the model to be interpreted as black box, we interpret the inner mechanisms of deep CNN models.  
\item The experiments are implemented in the TensorFlow\cite{b9} platform, which makes our model scalable to big datasets more easily. Scalability is an issue pointed out as future work in \cite{b6} and\cite{b7} but not realized yet. 
\end{itemize}

\section{RELATED WORK}
To resolve the problems for ``trusting a prediction" and ``trusting a model", two methods are proposed  in \cite{b6}  to explain individual predictions and understand a model's behavior respectively: Local Interpretable Model-agnostic Explanations (LIME) and Submodular Pick LIME (SP-LIME). The main idea for LIME is to use inherently interpretable models \textit{g} to interpret complex models \textit{f }locally. They designed an objective function to minimize the unfaithfulness (when \textit{g} is approximating \textit{f} in a local area) and the complexity of \textit{g}. Although it was stated in their paper that in the objective function \textit{g} could be any interpretable models, they set \textit{g} as sparse linear models in their paper.  Based on the individual explanations generated by LIME, they design an submodular pick algorithm: SP-LIME to explain the model as a whole by picking a number of representative and non-redundant instances.

It was suggested in \cite{b10} that coverage, precision and effort should be used to evaluate the results of the model interpretation. Although LIME achieves high precision and low effort, the coverage is not clear. In other words, LIME is able to explain why a specific prediction is made using the weights of the local model \textit{g}, but can't indicate to what local region the explanation is faithful. To solve this problem, the Anchor Local Interpretable Model-Agnostic Explanations method  (aLIME) was introduced in \cite{b10}. In aLIME,  the if-then rules are used instead of using the weights in a linear model to explain a specific prediction (as was executed in LIME).  The idea is based on the Decision Sets algorithm from \cite{b5}. These if-then rules are easy to comprehend and has good coverage.

It was pointed out that there is a trade off between interpretability and accuracy for machine learning algorithms\cite{b5}. In terms of inherently interpretable models, rule-based models, e.g. Decision Trees and Decision Lists are often preferred, as they can find a balance between these two factors. Decision lists are usually considered more interpretable than decision trees, as they use the if-then-else statements with a hierarchy structure. But this structure  reduces  to some extent the interpretability, as to interpret an additional rule all previous rules should be reasoned about. Also new rules down the list are applied to much narrow feature spaces, which makes the multi-class classification difficult where the minority classes deserves equally good rules. This motivates the proposal of the Decision Sets algorithm in\cite{b5}, which produces the isolated if-then rules, where each rule could be an independent prediction. To realize this, an objective function  takes into account  both interpretability (expressed by precision and recall of rules) and accuracy (expressed by size, length, cover and overlap). They showed that solving the objective function is a NP-hard problem, and finds  near-optimal solutions of it. However, Decision set' accuracy only approaches  random forest, and its expressive power just catches up with decision tree. 

Another model agnostic explanation approach is the Black Box Explanations through Transparent Approximations (BETA),  introduced in \cite{b7}. Different from LIME which aims for local interpretation, BETA is a framework which attempts to produce global interpretation for any classifier which are treated as black box classifiers. Based on their previous work on Decision Sets, the authors designed a framework with two level decision sets to taking into account fidelity (faithfulness to the black box model), unambiguity (single and deterministic explanations for each instance), interpretability (complexity minimized) and interactivity (user specified explorations of the feature's subspace). In this two level structure, the outer if-then rules are the ``\textit{neighborhood descriptors}" and the inner if-then rules are ``\textit{decision logic rules}" (how the black box model labels an instance under the outer if-then rules). Similar to \cite{b5}, an objective function is built and near-optimal solutions are found.

Two tools are introduced in \cite{b23} to provide intuitive understandings about the inner workings of deep neural networks (DNNs). The first tool directly plots the activations on each layer of a specific trained DNN. The second tool visualizes the features computed by each neurons on each layer of a DNN. Although the tools introduced in \cite{b23} also interpret DNNs, they are totally different approachs from ours. They visualize the values of the intermediate results directly within a DNNs while we visualize the behavior of a hidden layer via an inherently interpretable algorithm: Decision Tree. Therefore, they aims to improve the architecture of DNNs and provide inspirations for transfer learning, discriminative networks and generative models. And we focus on interpreting how the hidden layers classify test instances to ensure trust on a trained DNN.

\section{Methodology}
Our methodology could be classified as the \textit{post-hoc} interpretation \cite{b14}, where a trained model is given and the main task is to interpret it. This method is close to the second approach mentioned in the fourth paragraph of the introduction section, but is also different in many ways. First, the model to be interpreted here is not treated as a black box as we directly interpret the hidden layers of a deep CNN. Second, compared to LIME\cite{b6} which only has local interpretability, our method achieves global interpretability. Similar to LIME, we also provide qualitative interpretation with graphs to visualize the results. As our method interprets deep CNN via Meta-learning, we first briefly introduces deep CNN, meta-learning and then discuss our framework in details. 
\subsection{Deep Convolutional Neural Network}\label{AA}
\begin{figure}[htbp]
\centerline{\includegraphics[scale=0.4]{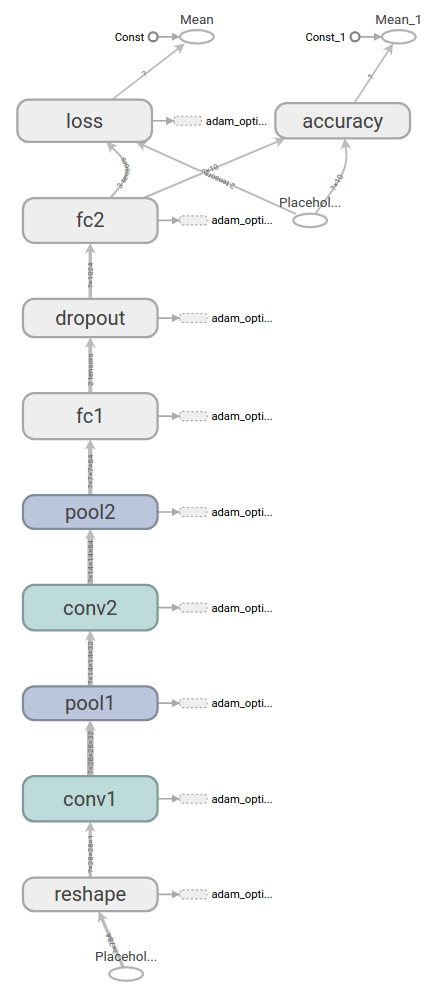}}
\caption{Structure of our deep CNN model generated by TensorFlow's TensorBoard .}
\label{fig_1}
\end{figure}

This section introduces the deep CNN model we are going to interpret. As we implement our program using TensorFlow, we use its TensorBoard function to draw the structure of the deep CNN we construct in Fig.~\ref{fig_1}. Deep CNN is now the most advanced machine learning algorithm for image classification. It takes advantage of the two-dimensional structure of the input images.  It uses a set of filters to filter the pixels of the raw input  images to generate higher level representations to be learnt by the model in order to improve the performance. 

There are three major components of deep CNN: convolutional layer, pooling layer and fully connected layer (same as in regular neural networks). A deep CNN model is usually a stack of these layers. In the convolutional layer, a filter is used to compute dot products between the pixels of the input image at specific position and the values of the filter, producing one single value in the output feature map. The convolution operation is completed after the filter is slided across the width and height of the input image. Following the convolutional layer, an activation function, often a rectified linear unit (ReLU) \cite{b15}, is applied to inject nonlinearities into the model and speed up the training process. Following ReLU is the pooling layer which is a non-linear down-sampling layer. A common algorithm for pooling is the max pooling algorithm. In this algorithm, each sub-region of the previous feature map is turned into a single maximum value in this region. Max pooling reduces computation and controls overfitting. In order to calculate the predicted class, after performing max pooling, the feature map needs to be flattened and feed into a fully connected layer. In the last layer: the output layer, a softmax classifier is applied for prediction.  

The structure of the deep CNN model we designed is illustrated in Fig.~\ref{fig_1}, ``Placeholder" represents for the interface to input the training data. ``Reshape" is needed first to convert the input one-dimensional image data into two dimensional data. In our experiment, we use the MNIST dataset\cite{b16}. 
The 784 input  features are converted into a two-dimensional  $28 \times 28$  image. Our model has two series of a convolutional layer  followed by a pooling layer: ``conv1"-``pool1"-``conv2"-``pool2", which  are followed by one fully connected layer ``fc1''.  As a fully connected network is susceptible to suffer from overfitting, the ``dropout" operation\cite{b17} applied after ``fc1'' aims to reduce it. In this operation, a  probability parameter \textit{p} is set to keep a specific neuron with probability \textit{p} (or drop it with probability \textit{1-p}). The ``Adam optimizer"\cite{b18}, rather than a standard Stochastic Gradient Descent optimizer is used  to train the model via modifying the variables and reducing the loss. ``fc2'' is the output layer with 10 neurons: each represents one of the classes: 0-9. 

\subsection{Meta-learning}
Meta-learning is an ensemble learning method which learns from the results of the base classifiers. It has a two-level structure, where the algorithms used in the first level are called base-learners and the algorithm in the second level is called the meta-learner. The base-learners are trained on the original training data. The meta-learner is trained by the predictions of the base classifiers and the true class of the original training data. When training the meta-learner, the ``Class-combiner'' strategy \cite{b13} is applied here, where the predictions  includes just the predicted class (instead of all classes, as in the ``Binary-class-combiner"). 

\begin{figure}[htbp]
\centerline{\includegraphics[scale=0.35]{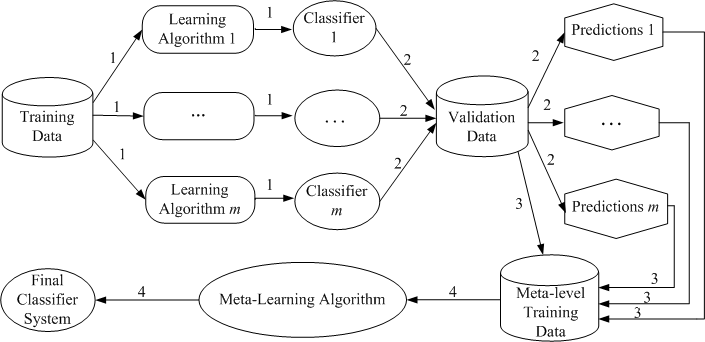}}
\caption{Meta-learning training process.}
\label{fig_2}
\end{figure}

To understand the meta-learning algorithm intuitively, Fig.~\ref{fig_2} 
illustrates a simplified training process for meta-learning \cite{b19}. The numbers 1, 2, 3, 4 represent the four steps of training. In the 1st step, the base learning algorithms 1 to \textit{m} are trained on the training data. In the 2nd step, a validation dataset is used to test the trained classifiers 1 to \textit{m}. In the 3rd step, the predictions generated in step 2 and the true labels of the validation dataset are  used to train a meta-learner. Finally, in the 4th step, a meta-classifier is produced and the whole meta-learning training process is completed.

\begin{figure}[htbp]
\centerline{\includegraphics[scale=0.33]{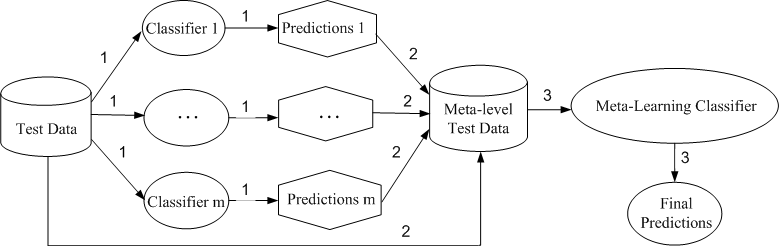}}
\caption{Meta-learning test process.}
\label{fig_3}
\end{figure}

Once the training process is accomplished, the test process is much easier to execute. Fig.~\ref{fig_3} presents a simplified test process \cite{b19}. In the 1st step, the test data is applied on the base classifiers to generate predictions which combined with the true labels of the test data comprises the meta-level test data in 2nd step. In the 3rd step, the final predictions are generated by testing the meta-level classifier with the predictions in the 2nd step and the accuracy could be calculated.

\subsection{Framework}
Our framework is named as CNN-INTE which stands for Convolutional Neural Network Interpretation. It is similar to meta-learning, but different in a few ways. In this work, we interpret the first fully connected layer ``fc1'' of the deep CNN model illustrated in Fig.~\ref{fig_1}. 

\begin{figure}[htbp]
\centerline{\includegraphics[scale=0.37]{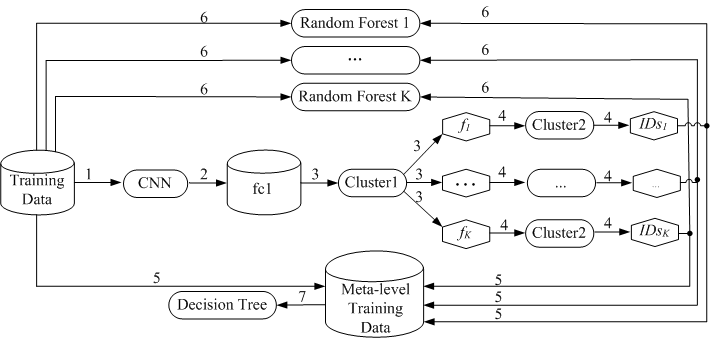}}
\caption{CNN-INTE training process.}
\label{fig_4}
\end{figure}

The training process is shown in Fig.~\ref{fig_4}. In the 1st step, the original training data is used to train a CNN model. In the 2nd step, the parameters generated in the 1st step are used to calculate the values for the activations of the first fully connected layer: fc1. In the 3rd step, a clustering algorithm is used to cluster the data generated in step 2 into a number of groups which we define as factors henceforth. In the 4th step, the data belonging to each of the factors are clustered again generating a number of clusters each assigned a unique  ID. In the 5th step, these IDs are grouped together as the training features in the meta-level, using the labels of the original training data as label for the meta-learner. In the 6th step, the features of the original training data and the IDs (set as labels) in step 4 are used to train a number of random forests \cite{b21}.

Now we discuss the training process in more details. Assume the training data $T$ has $N$ numbers of instances and layer ``fc1'' has $H$ neurons. The labels of the training data are $T_y=\left\lbrace l_1, l_2, ..., l_N\right\rbrace$. Once the deep CNN model is trained, for each training instance $t_i$, we calculate the activations at each hidden neuron on this layer. Hence, we obtain a matrix $S$ with size $H\times N$. To construct the meta-level training data, we use a clustering algorithm to cluster $S$ along the hidden layer axis into several factors $F=\left\lbrace f_1, f_2, ..., f_K\right\rbrace$. How to set the value of $K$ is tricky. In our experiments, this value is determined as the one which produces the best accuracy performance for the meta-learning algorithm. We also discussed how to avoid this problem in section \Rmnum{5}.  Then within each of the factors, we cluster the data again, this time along the axis of the instances. The clustering results are the IDs each instance belongs to. For instance, if the number of clusters is 10, after the second level clustering  each instance will  have an ID number between 0-9. All the IDs combined with the true labels of the training data builds up the meta-level training data. 

To present the technical  details of the CNN-INTE training process, we provide the pseudo code in Algorithm \ref{alg_1}. Line 1-3 is the initialization of the algorithm. In line 4, the activations $S$ are clustered into $K$ factors, where $K$ is the number of clusters set in the first level clustering algorithm $C^l$. In lines 5-7 the same clustering algorithm $C^l$ is applied on all the factors to generate $K$ sets of ID numbers. Lines 8-9 uses the generated ID numbers and the true labels of the original training data to train the meta-learner: $C^m$. Till now, the training process is not done yet. We still need to generate the base models to be used in the test process. Lines 10-12 uses the features of the original training data and the ID numbers to train $K$ base models. The output of the training process would be the meta-lever classifier: $\tilde{M}$ and $K$ base models: $B=\left\lbrace M_1,M_2,\cdots,M_K \right\rbrace$. 

\begin{figure}[htbp]
\centerline{\includegraphics[scale=0.33]{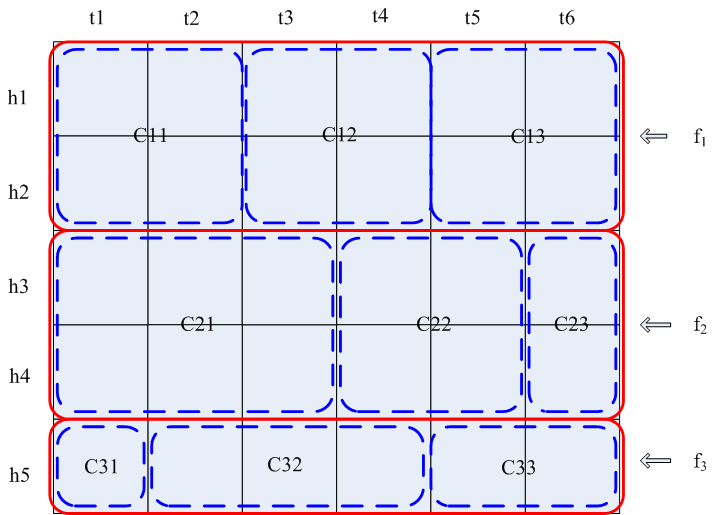}}
\caption{Toy example for the generation of Meta-level training data.}
\label{fig_5}
\end{figure}

\begin{algorithm}
\caption{CNN-INTE Training Process}\label{alg_1}
\begin{algorithmic}[1]
\State Input: activations: $S$; training data: $T$; Meta learning algorithm: $C^m$; Clustering algorithm: $C^l$; Base learning algorithms: $\left\lbrace C_1,C_2,\cdots,C_K \right\rbrace $
\State $E=\varnothing$
\State $S_{cv}=\varnothing$
\State $\left\lbrace f_1, f_2,\cdots,f_K \right\rbrace=C^l(S)$
\For{$k=1\cdots K$}
\State $IDs_{k}=C^l(f_k)$
\EndFor
\State $S_{cv}=\left\lbrace IDs_{1},IDs_{2},\cdots,IDs_{K}, T_y \right\rbrace $
\State $\tilde{M}=C^m(S_{cv})$
\For{$k=1\cdots K$}
\State $M_k=C_k(T_x, IDs_k )$
\EndFor
\State $E=(\left\lbrace M_1,M_2,\cdots,M_K \right\rbrace,\tilde{M})$
\State Output: Ensemble $E$ 
\end{algorithmic}
\end{algorithm}
Fig.~\ref{fig_5} is a toy example that illustrates the above process. In this example, there are 5 hidden neurons and 6 training instances. We set the number of clusters for both the first and second level clustering as 3. Hence,  the matrix  $S$ with size  $ 5\times 6$ is first clustered into 3 factors  $\left\lbrace f_1, f_2, f_3\right\rbrace$ horizontally. For each factor, the activations are again clustered into three clusters vertically, e.g. $F_1$ is clustered into $\left\lbrace C_{11}, C_{12}, C_{13}\right\rbrace$. If we set the ID numbers for these cluster as  $\left\lbrace 0, 1, 2\right\rbrace$, then the corresponding ID numbers for $t1$ to $t6$ in factor $f_1$ according to Fig.~\ref{fig_5} are $\left\lbrace 0, 0, 1, 1, 2, 2\right\rbrace$. Hence, the meta-level training features are expressed as 
\[
 \begin{bmatrix}
  0 & 0 & 1 & 1 & 2 & 2 \\
  0 & 0 & 0& 1 & 1 & 2\\
  0 & 1 & 1 & 1 & 2 & 2
 \end{bmatrix}
\] 
This data combined with the corresponding training labels of the original training data is used to train the meta-learner. Here the meta-learner we used is the Decision Tree \cite{b3}, an inherently interpretable algorithm. Its tree structure provides an excellent visual explanation of the predictions. 

\begin{figure}[htbp]
\centerline{\includegraphics[scale=0.33]{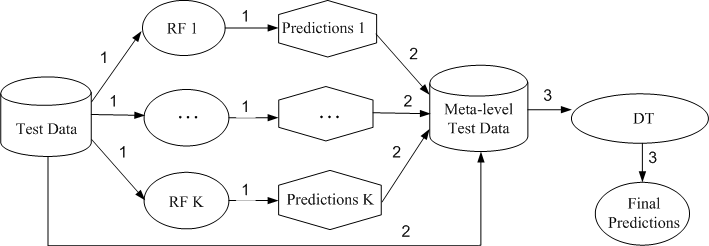}}
\caption{CNN-INTE test process.}
\label{fig_6}
\end{figure}

The test process of the meta-model is exactly the same as the meta-learning test process, which is shown in Fig.~\ref{fig_6}. 
In the test process, we use the original test data to test the base classifiers generated in the meta-level training process to obtain the meta-level test data's features. The base-learner we applied is random forest. The number of base models is equal to the number of factors. Hence, we have $K$ base models: $B=\left\lbrace M_1, M_2, ..., M_K\right\rbrace$. In the toy example, there are three factors which lead to three base models. The training data for the first base model corresponding to $f_1$ are
\[
 \begin{array}{c|c}
  (t1-l1) & 0\\
  \hline
  (t2-l2) & 0\\
  \hline
  (t3-l3) & 1\\
  \hline
  (t4-l4) & 1\\
  \hline
  (t5-l5) & 2\\
  \hline
  (t6-l6) & 2\\
 \end{array}
\]

Here $t_i-l_i$ represents for the features of the original training instance $i$. Once we obtain the $K$ base models,  we can use the original test data to test them to produce the meta-level test data. These data are then feed into the trained decision tree model to interpret individual test predictions. 

\section{Experiments}
The dataset we use is the MNIST database of handwritten digits from 0 to 9\cite{b16}. We extracted 55,000 examples (the original dataset has 60,000 examples for training) as the training data and 10,000 examples as the test data. Each of the examples represents for the $28\times 28$ images with pixels flattened as 784 features. The experiments are performed on the TensorFlow platform. 
\subsection{Experimental Setup}\label{AA}
First of all, we need to train a nice deep CNN model. We first reshape the input training data into 55000 images each with size $28\times 28$. Training all the data on every epoch is expensive, which requires lots of resources of the computer and may lead to the termination of the program. Here we apply stochastic training: on the first epoch, we select a mini-batch of the training data and perform optimization on this batch; once we loop through all the batches, we randomize the training data and start a new epoch. In our experiment, we set the epoch $e=1000$, batch size $b=50$. Stochastic training is cheap and achieves similar performance to using the whole training data in every epoch. For each mini-batch, in the first convolutional layer, we apply 32 filters (or kernels) each with size  $5\times 5$, which generates 32 feature maps. In the first pooling layer we apply filters with size $2\times2$. The stride size is set as 2. The second convolutional layer use 64 filters with the same size as the first convolutional layer. The second pooling layer has the same parameters as the previous one. Immediately after this pooling layer is the first fully connected layer: fc1. We set the number of neurons for this layer as 128. To reduce overfitting we also set the dropout \cite{b17} parameter $d=0.5$, which means a neuron's output has 50\% probability to be dropped. The last layer is the second fully connected layer (or the ``readout layer''), which has 10 neurons with each neuron outputs the probability of the corresponding digits 0-9.   The test accuracy of this trained CNN model on the test data is 93.9\%. 

Now comes the key part for setting up interpretation. we define interpretability of a model as the ability to provide visual or textual presentation of the connections between input features and the output predictions. We first feed the trained fully connected layer $fc1$ with the original training data, which would produce a data $S$ with size of $128\times 55000$. We then cluster $S$ into several factors. The clustering algorithm we applied is the k-means algorithm \cite{b20}.  The number of factors is equal to the number of clusters which we set as 8 in this level. Hence, $S$ is now turned into a list $F=\left\lbrace f_1, f_2, ..., f_8\right\rbrace$ with size $8\times55000$ having each row representing the data belonging to each factor. In the second level clustering, for each factor in $F$ we use the k-means algorithm again to cluster them into a number of clusters. We set the number as 10 in our experiment as the number of classes for the original training data is 10. Hence each cluster will be assigned a unique ID number between 0 and 9. Then we use the IDs belonging to each training instances and the true labels of the original training data to train a decision tree algorithm. Due to the limitation of the space, we are unable to show the structure of the trained decision tree here. We set the maximum depth of the decision tree as 5. Although deeper decision tree would generate better accuracy, it makes it harder to interpret with too many tree levels. 

To obtain the test data for decision tree, we first use the original training data's feature as features and the IDs for each factor in $F$ as labels to train the corresponding random forest algorithm\cite{b21}, generating 8 base models. For random forest, we set the number of trees as 20 and the maximum nodes as 2000. Finally we use the original test data to test the 8 trained base models. The generated predictions become the features of meta-level test data with sizes of  $10000\times8$. Using the meta-level test data on the trained decision tree produces an accuracy of 92.8\% with tree depth=5. This value is comparable to the test accuracy on the trained  deep CNN model: 93.9\%. It should be noted that the decision tree's accuracy could be further improved by increasing the depth of tree and tuning other related parameters. 

\subsection{Experimental Results}
To interpret the deep CNN model's behavior on the test data, we intend to use diagrams generated by our tool: CNN-INTE to examine individual predictions on the test data. Hence, we provide qualitative interpretations visually. We arbitrarily selected two test instances that were correctly classified by the decision tree and one test instance that was wrongly classified. It should be noted that this tool could be used on any test instances globally and not just limited to the three cases we provide. The details of the selected test instances are shown in Table~\ref{tab1}. Here ``$f_0$-$f_7$'' represents  the features of the meta-level test data, ``label" is the test label in the original test data, ``pred'' is the prediction generated by the decision tree on the meta-level test data. ``True1'' and ``True2'' represents for the two correctly classified instances and ``Wrong1'' is the wrongly classified instance.

\begin{figure*}[h]
  \includegraphics[width=\textwidth]{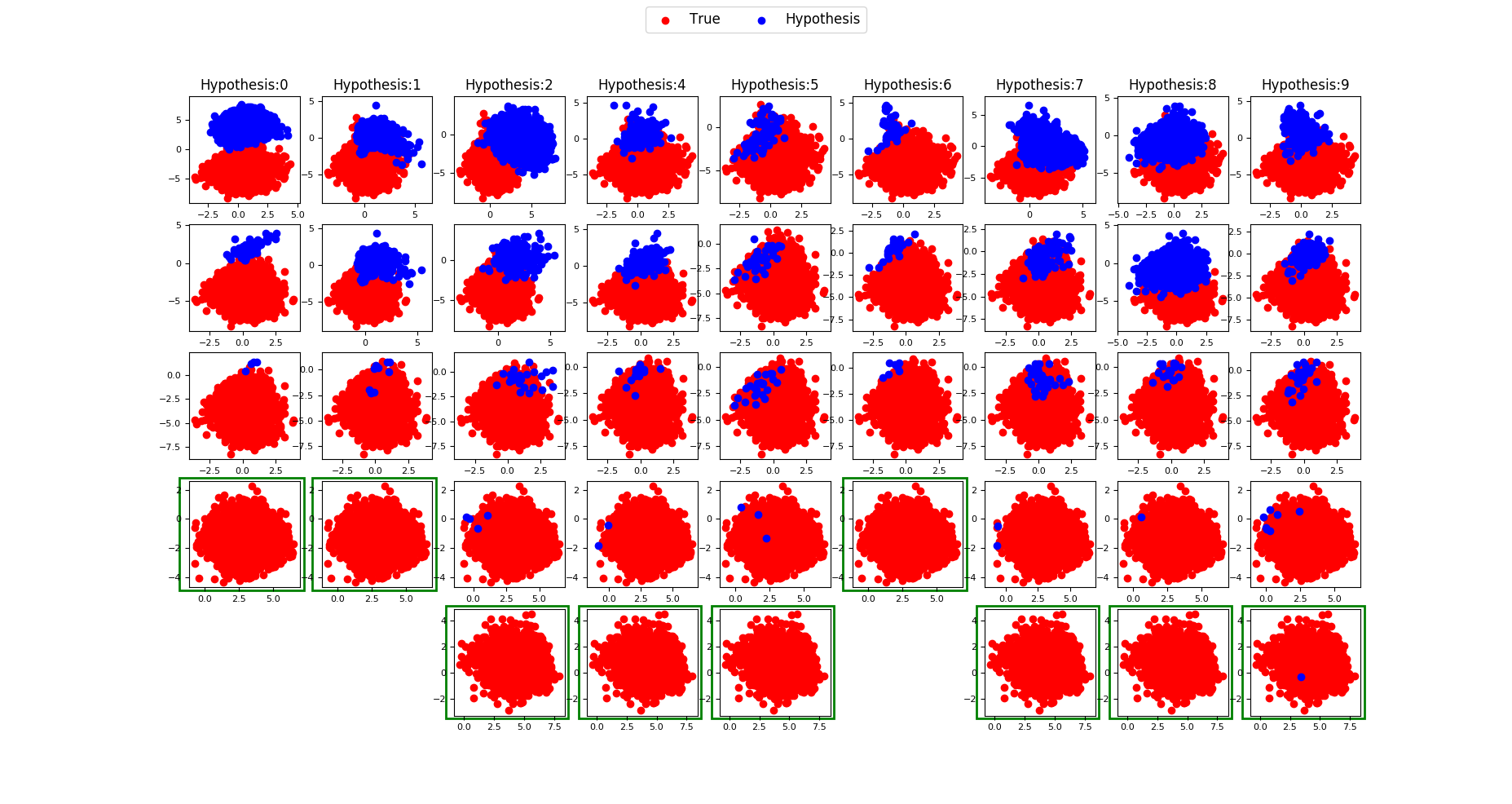}
  \caption{Example of a correctly classified test instance: True1.}
  \label{fig_7}
\end{figure*}

In order to examine the classification process visually, we check each feature values according to the trained structure of the decision tree and plot the graphs of the activations corresponding to the true label and the hypothesis. The interpretation result for instance ``True1'' is shown in Fig.~\ref{fig_7}. As the true label for this instance is 3, all other classes could be regarded as hypothesis and this is why there are no graphs for ``Hypothesis: 3" in Fig.~\ref{fig_7}. Each row represents  the examination of the feature values corresponding to different factors in different levels of the trained decision tree, e.g. the first row represents  the root level of the decision tree. Since we set the depth of the decision tree as 5, there are 5 rows in all. Each column stands for the query of if the test instance belongs to the corresponding hypothesis over the nodes visited. 

\begin{figure*}[h]
  \includegraphics[width=\textwidth]{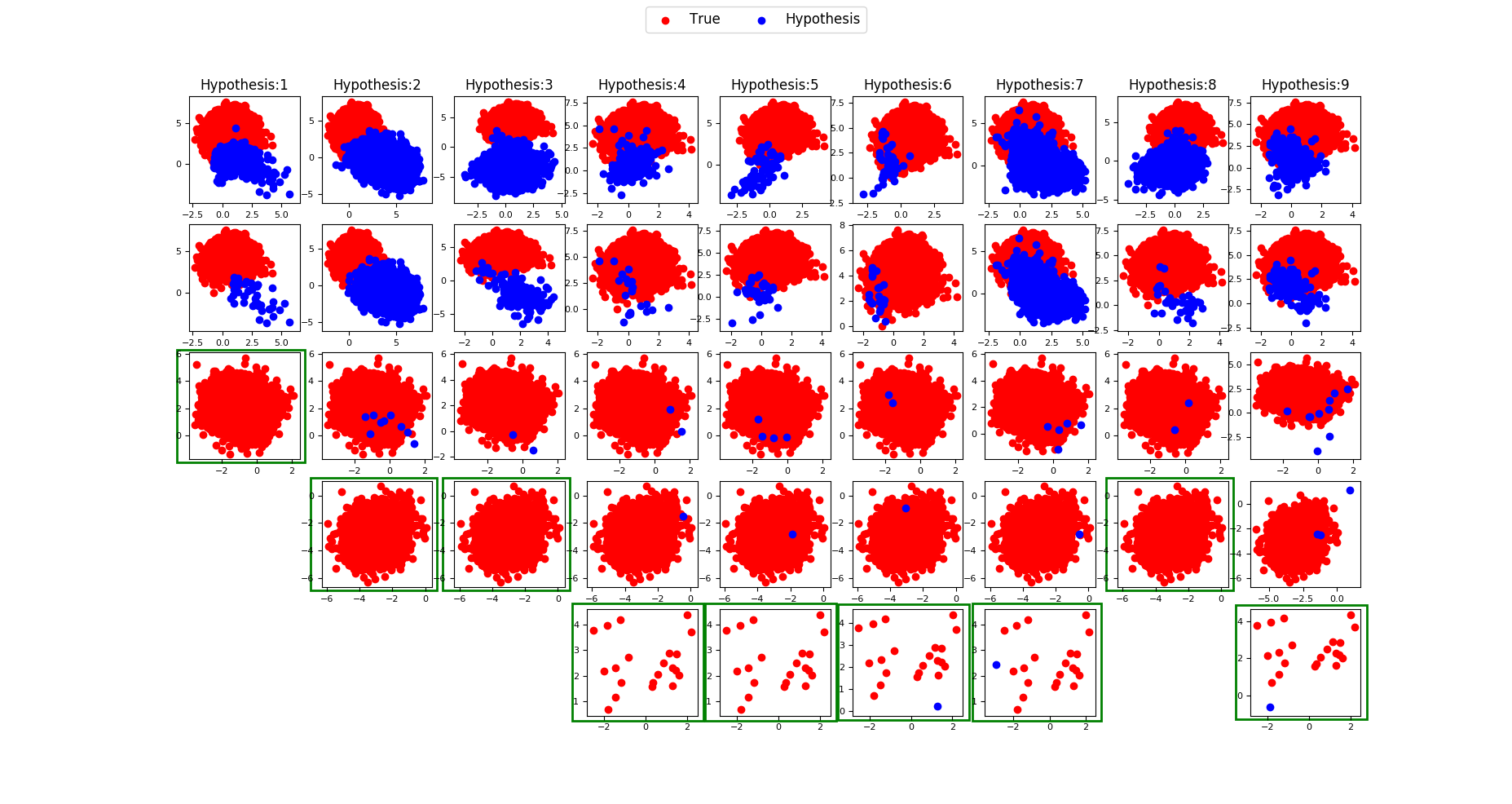}
  \caption{Example of a correctly classified test instance: True2.}
  \label{fig_8}
\end{figure*}

\begin{figure*}[h]
  \includegraphics[width=\textwidth]{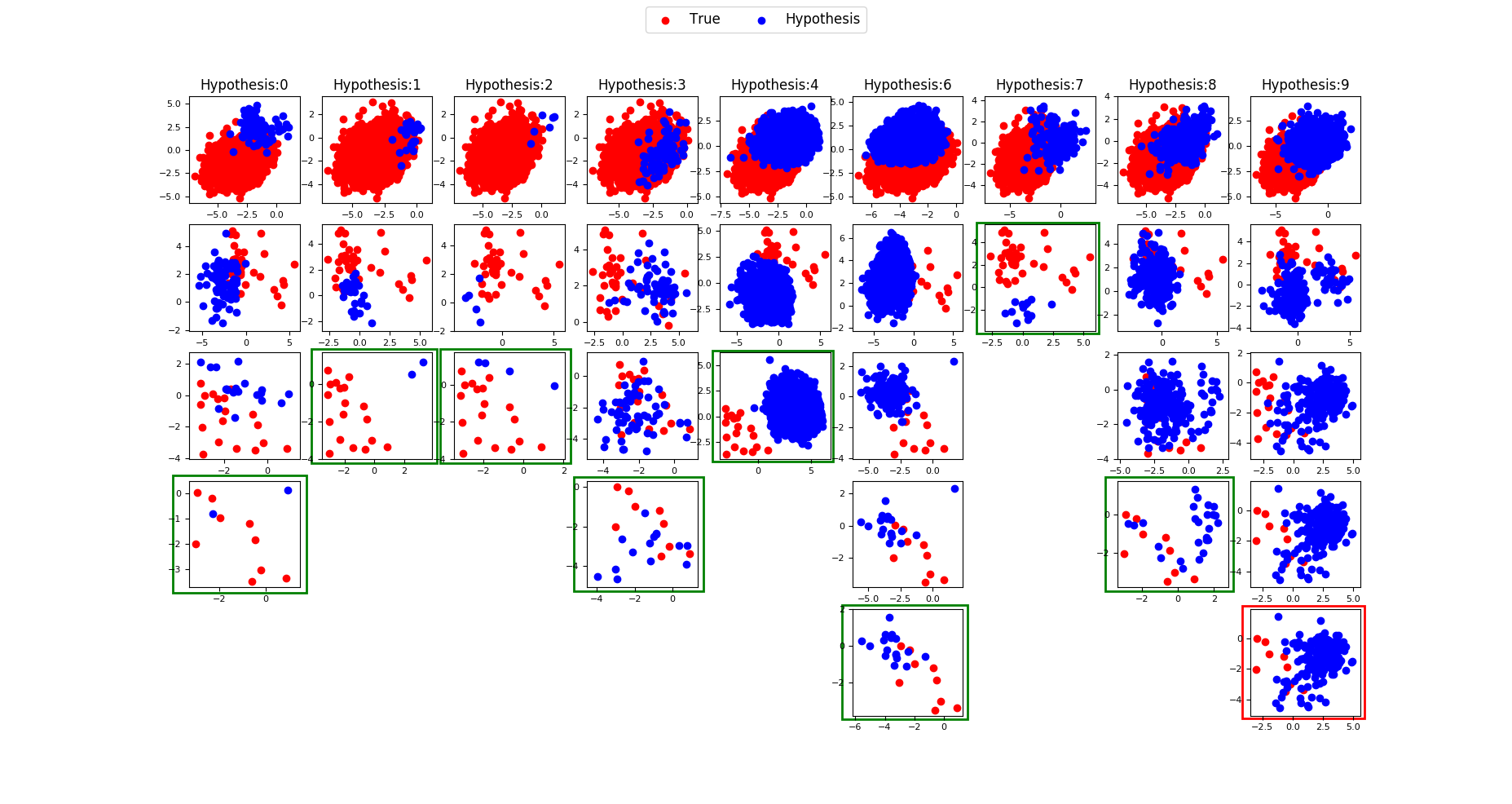}
  \caption{Example of a wrongly classified test instance: Wrong1.}
  \label{fig_9}
\end{figure*}

Take the column of ``Hypothesis:0'' as an example,  the goal is to find if the label of the test instance is 0. In the 1st row we extract the activations corresponding to ``$f_6$''  which satisfies the condition that $f_6\leqslant 4.5$ (this is determined by the trained decision tree) and draw a graph between activations that belongs to label=0 (hypothesis) and label=3 (true). Then we check the graph to evaluate if the data corresponding to the true class could be separated from the hypothesis. The answer is no because the hypothesis represented as blue points overlaps with the true class shown as red points. Hence, we need to query the trained decision tree further. The values of the factors we need to check is: $f_6\leqslant 2.5$ for 2nd row;  $f_6\leqslant 0.5$ for 3rd row; $f_7\leqslant 0.5$ for 4th row; $f_1\leqslant 0.5$ for 5th row. In this process, we noticed that in the 4th row the true class and the hypothesis class are successfully separated as only the red points corresponding to the true label are left. Therefore, we don't need to examine further and that's why the graph for the 5th row is not displayed. We highlight the graph with green rectangles if the final results are separable and red vice versa. The same idea is applied on other hypothesis. We also draw the graphs for instances ``True2'' and ``Wrong1'' in Fig.~\ref{fig_8} and Fig.~\ref{fig_9} respectively.
 \begin{table}[htbp]
\caption{Instances Selected From the Meta-level Test Data}
\begin{center}
\begin{tabular}{|c|c|c|c|c|c|c|c|c|c|c|}
\hline
\textbf{}&\multicolumn{10}{|c|}{\textbf{Features and labels}} \\
\cline{2-11} 
\textbf{} & \textbf{$f_0$} & \textbf{$f_1$} & \textbf{$f_2$} & \textbf{$f_3$} & \textbf{$f_4$} & \textbf{$f_5$} & \textbf{$f_6$} & \textbf{$f_7$} & \textbf{\textit{label}}& \textbf{\textit{pred}}\\
\hline
True1&4 &0 &7 &7 &0 &0 &0 &0 &3 &3\\
\hline
True2&5 &0 &0 &5 &9 &5 &3 &6 &0 &0\\
\hline
Wrong1&5 &6 &7 &9 &6 &4 &7 &9 &5 &9\\
\hline
\end{tabular}
\label{tab1}
\end{center}
\end{table}


\section{Conclusion and future work}
In this work, we present an interpretation tool CNN-INTE, which interprets a hidden layer of a deep CNN model: to find out how the learned hidden layer classifies new test instances. Although we just show the results for the first fully connected layer before the read-out layer, the approach could be applied on any hidden layers. The interpretation is realized by finding the relationships between the original training data and the trained hidden layer ``fc1'' via meta-learning. We used two-level k-means clustering algorithm to find the meta-level training data and random forests as base models for generating meta-level test data. The visual results generated by our program clearly indicate why a test instance is truly or wrongly classified by checking if there are any overlaps of the corresponding activations. 
For future work, we plan to initiate quantification of the interpreted results. In our experiments, one of the things we find tricky is the setting of the number of clusters for the k-means algorithm. In the future, we plan to replace the k-means algorithm with DBSCAN \cite{b22} which doesn't need specifying the number of clusters.  As stated in \cite{b5}, ``decision sets" seems to be a better option than decision tree as a inherently interpretable algorithm, so we also plan to replace decision tree with decision sets. Last but not least, it would be quite meaningful to apply this tool on real world applications with more complex data where interpretations are demanded either between the training data and the hidden layer or between the hidden layer and the predictions.

\section*{Acknowledgment}

The authors acknowledge the support of the Province of Nova Scotia, of Dalhousie University, and of the  the Natural Sciences and Engineering Research Council of Canada under the CREATE program grant.

\end{document}